\documentclass[conference]{IEEEtran}
\ifCLASSINFOpdf
  \usepackage[pdftex]{graphicx}
  \graphicspath{{./images/}}
  \DeclareGraphicsExtensions{.pdf,.jpeg,.png}
\else
\fi
\begin{document}
%
\title{A Multiple Radar Approach for Automatic Target Recognition of Aircraft using Inverse Synthetic Aperture Radar}

\author{\IEEEauthorblockN{
    Carlos Pena-Caballero\IEEEauthorrefmark{1},
        Elifaleth Cantu\IEEEauthorrefmark{1}, 
        Jesus Rodriguez\IEEEauthorrefmark{1}\\
        Adolfo Gonzales\IEEEauthorrefmark{1},
        Osvaldo Castellanos\IEEEauthorrefmark{1},
        Angel Cantu\IEEEauthorrefmark{1}\\
        Megan Strait\IEEEauthorrefmark{1},
        Jae Son\IEEEauthorrefmark{2} and
        Dongchul Kim\IEEEauthorrefmark{1}}
    \IEEEauthorblockA{\IEEEauthorrefmark{1}Department of Computer Science\\
    }
    \IEEEauthorblockA{\IEEEauthorrefmark{2}Department of Electrical Engineering\\
        University of Texas Rio Grande Valley,
        Edinburg, Texas 30332--0250\\
        Email: dongchul.kim@utrgv.edu}
}


%


\maketitle

\begin{abstract}
Following the recent advancements in radar technologies, research on automatic target recognition using Inverse Synthetic Aperture Radar (ISAR) has correspondingly seen more attention and activity.  ISAR automatic target recognition researchers aim to fully automate recognition and classification of military vehicles, but because radar images often do not present a clear image of what they detect, it is considered a challenging process to do this. Here we present a novel approach to fully automate a system with Convolutional Neural Networks (CNNs) that results in better target recognition and requires less training time. Specifically, we developed a simulator to generate images with complex values to train our CNN. The simulator is capable of accurately replicating real ISAR configurations and thus can be used to determine the optimal number of radars needed to detect and classify targets. Testing with seven distinct targets, we achieve higher recognition accuracy while reducing the time constraints that the training and testing processes traditionally entail.
\end{abstract}


%
\IEEEpeerreviewmaketitle

\section{Introduction}
\begin{figure}[h!]
  \includegraphics[width=\linewidth]{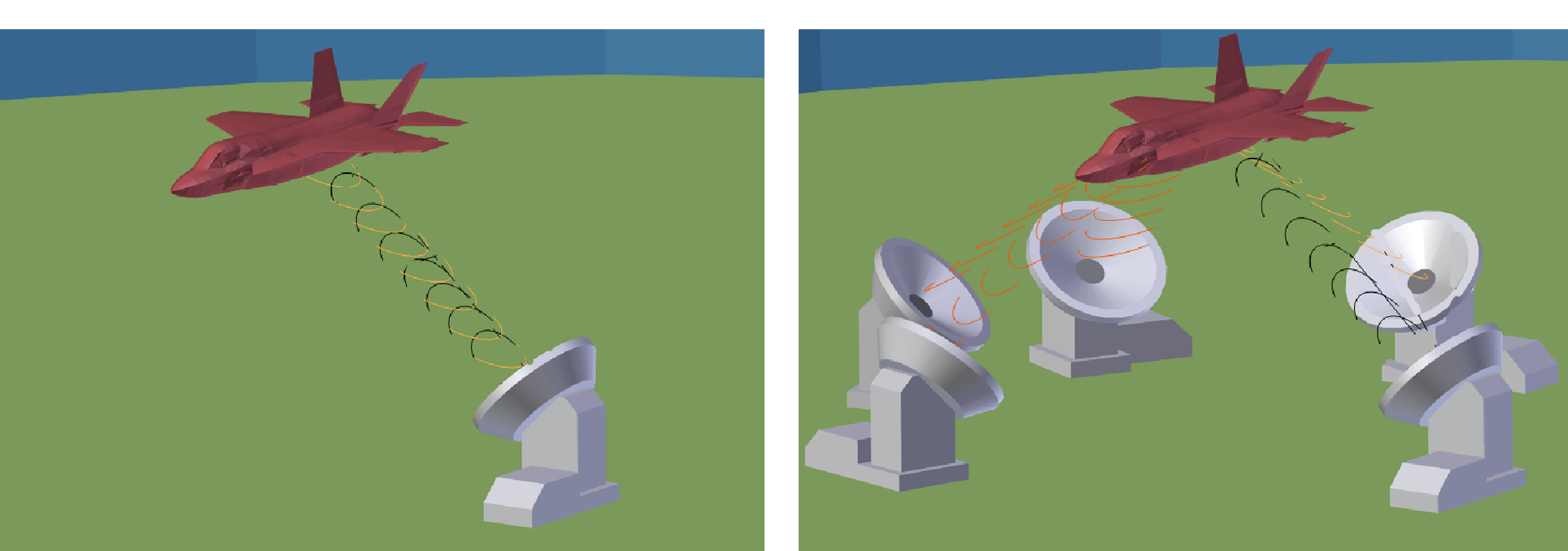}
  \caption{a) Mono-static radar, b) Multiple Mono-static radar}
  \label{radars}
\end{figure}

Along with the improvement of radar technologies, as well as high demands in target identification in radar application, the Synthetic Aperture Radar (SAR) and ISAR automatic target recognition are powerful techniques to generate high-resolution images two-dimensional images virtually in any type of weather conditions and lighting.\\
Thanks to SAR/ISAR techniques we are able to obtain a clear image of any target and classify it properly, but doing so takes considerable time to analyze manually; thus automatic target recognition aims to automate this process and let computer analyze the radar-generated pictures in order to process all the data in real time as fast and efficiently possible. The Moving and Stationary Target Acquisition and Recognition (MSTAR) program is a state-of-the-art model-based vision approach to SAR automatic target recognition \cite{9}; the MSTAR program has been used by many researchers to create and validate their algorithms, in this paper we created an internal simulation software called "RadarPixel", which we used to generate all the necessary data to test our approach. Using the MSTAR dataset of a real Slicy we created a three-dimensional virtual model based on the specifications made by Wong \cite{4}, and successfully validated our simulator comparing visually the resulting images from our simulator versus the MSTAR dataset.\\
In this paper we present a novel approach to process and classify military aircraft in real time, which will effectively eliminate the necessity of human operator sift through all the generated images of the radar; our approach will consist in a multiple array of radars strategically place in an area that will help maximize the area of cover for target recognition giving almost a full 360 degrees of coverage around any one target, thus resulting in higher accuracies and faster classification, even when the weather conditions are not favorable (i.e. noise in the image).\\
Normally the ISAR methods include only one radar sending and receiving the electromagnetic waves bouncing off of a target (see Figure ~\ref{radars}.a) this approach is called Mono-static radar, but our approach includes an array of strategically placed radars, where one send signals to the target and the rest receive the signal, in Figure ~\ref{radars}.b we have mocked-up a possible scenario of this approach. We call our approach Multiple Mono-static radar. 
The arrangement of this paper is as follows: In Section II we will explore a few related works in the SAR automatic target recognition research field and look at their accomplishments and contributions to the field, in Section III we will explain the simulator used to get the SAR images for this experiments and how it was validated, as well as the necessary parameters to replicate our data generation. During Section IV we will dive into the architecture for our CNN and the resulting dataset generated from the simulator, in Section V we will see the experimental results of our tests and in Section VI and VII is for conclusions and discussion.\\

\begin{figure*}
 \centering \includegraphics[width=\linewidth]{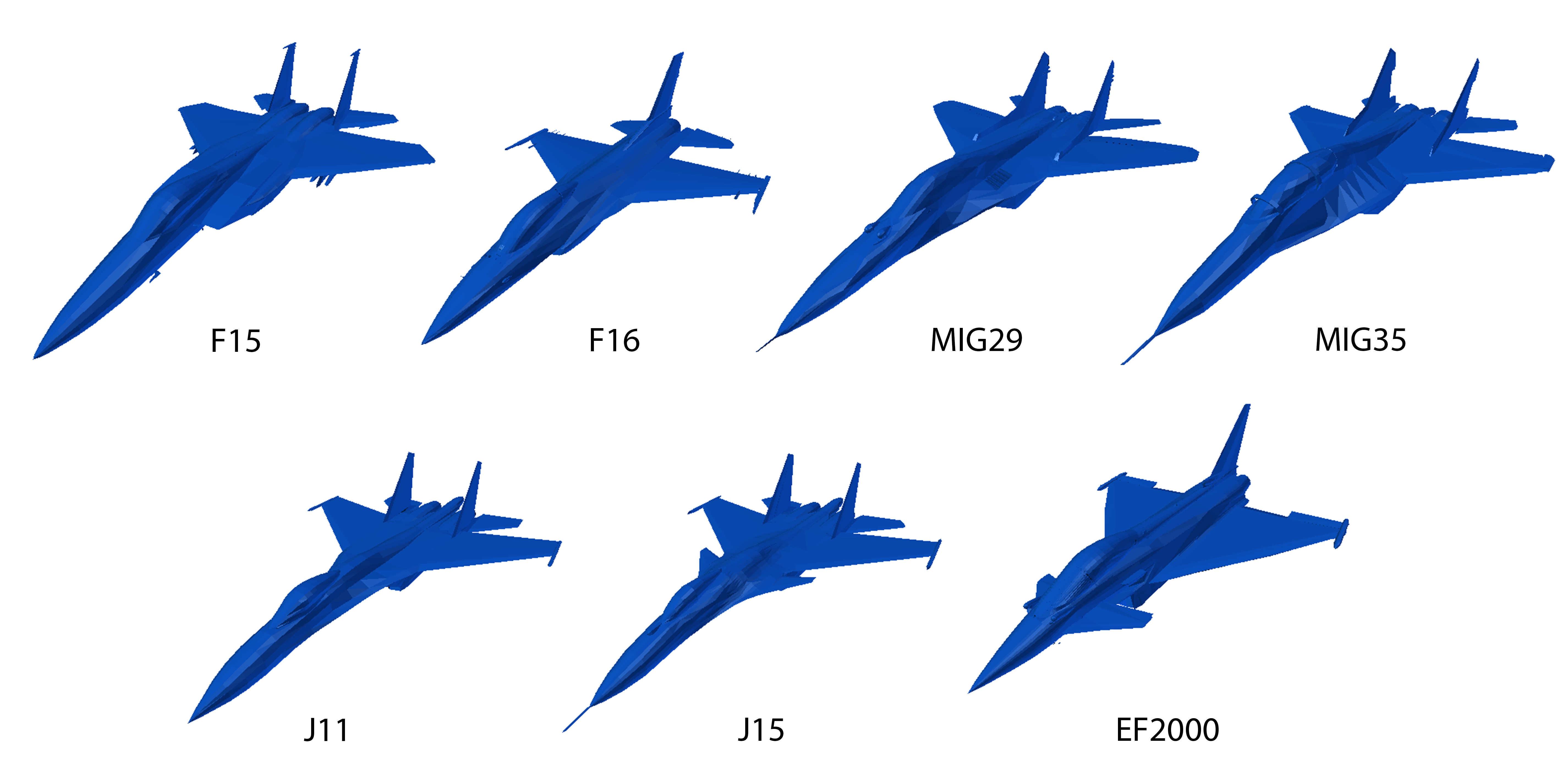}
  \caption{Aircraft models used during the simulations}
  \label{aircraft}
\end{figure*}

\section{Related works}

As an active research topic and it's extensively applications to broader problems, the SAR/ISAR automatic target recognition techniques usage have been applied in the fields regarding surveillance, homeland security, and military tasks (e.g., \cite{10}-\cite{13}). Novak et al. \cite{13} used a simple mean square error classifier (MSE) to accomplish accuracies of 66.2\% and 77.4\% with 20 and 10 target classes respectively on the MSTAR dataset.
Bhanu et al. [10], presents a variety of paper in the field of automatic target detection and recognition; the authors made specific groups to categorize the different techniques used in the automatic target detection and recognition, the first group uses “multiresolution processing for clutter modeling, target detection, and recognition,” the second group relates to  “physics-based processing for target detection, recognition, and change detection,” the next one use “geometrical approaches to target detection and recognition,” and the last group is“model-based processing of image sequences for target motion detection, recognition, tracking and change detection for wide area surveillance.”
Martone, Innocenti and Ranney \cite{11}, proposed a system to detect and track moving personnel inside wood and cinderblock structures, using an automatic target detection algorithm, a constant false alarm rate approach, and k-means clustering they were able to achieve their goal.
Past work in SAR automatic target recognition involved the use of SVM such as Gorovyi and Sharapov \cite{6}, where an SVM achieved an accuracy of 90.7\% on the MSTAR data set. Wagner \cite{7} extended their SVMs with convolutional layers improving performance and achieving an accuracy of 99.5\% for forced classification on the MSTAR data set. Zhong and Ettinger\cite{8} use conventional CNNs to reach accuracies that range from 99.0\% to 99.5\% on the MSTAR data set; this suggests that convolutions work well on SAR in order to improve the performance of SAR target recognition on the MSTAR data set compared to conventional SVMs. 

Although Wanger \cite{7} and Zhong and Ettinger \cite{8} both use CNNs to train and test their databases we utilize an unusual approach that leads a better performance and faster training times since our CNN uses the complex values generated by the simulation itself to train the network. We utilize a multi-radar approach to increase the accuracy of the training and testing processes, thus resulting in higher accuracies than the other papers working on SAR/ISAR automatic target recognition. The only flaw with our approach would be that it is a very time-consuming job to generate the data since we need to gather an enormous amount of data to get more accurate results from the network.\\ 

\section{Simulation}
\subsection{RadarPixel}

\begin{figure}
  \includegraphics[width=\linewidth]{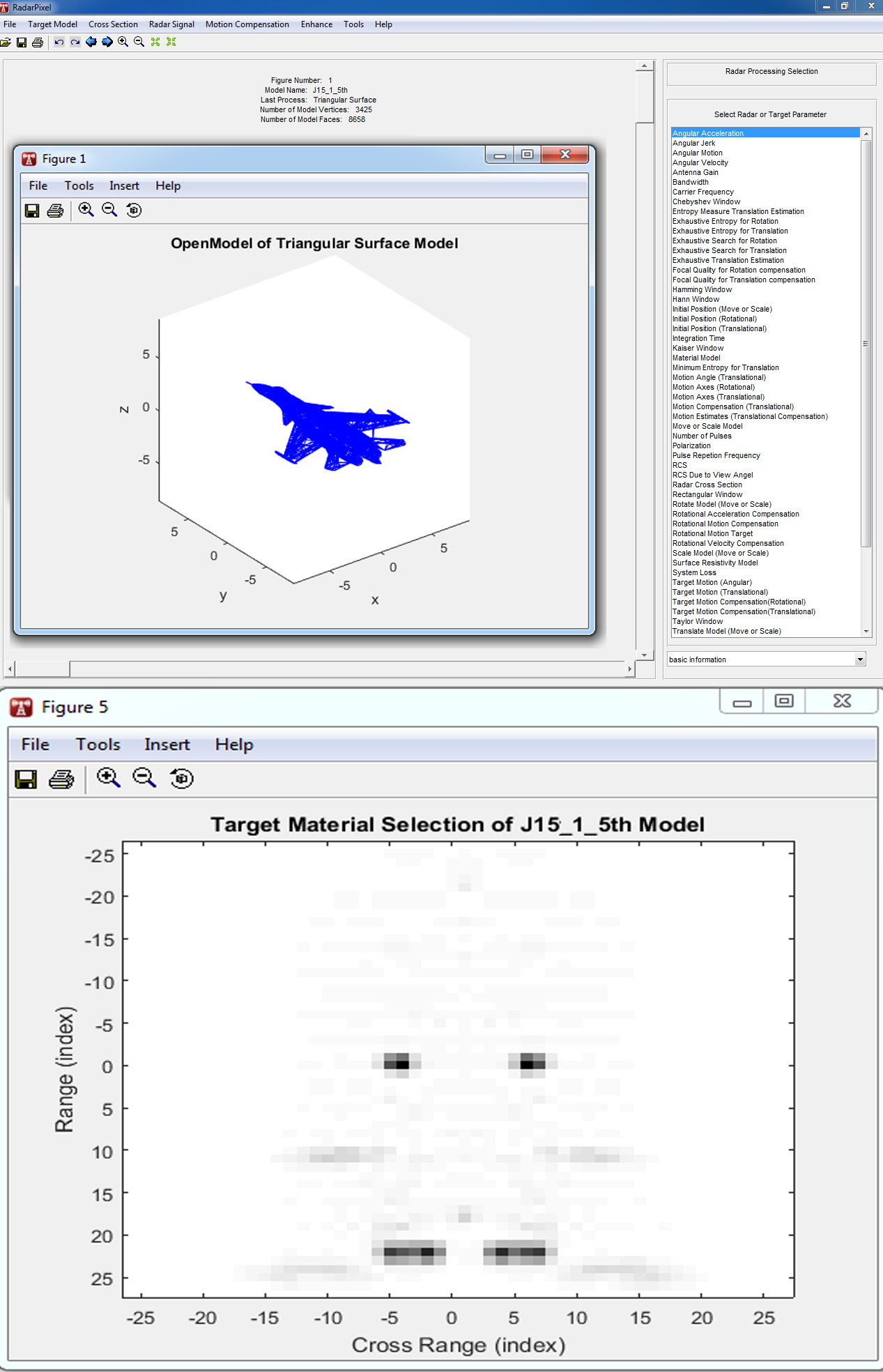}
  \caption{a) RadarPixel plane model, b) RadarPixel simulated SAR image}
  \label{radarPixel}
\end{figure}

RadarPixel \cite{27} is a simulation software developed by us able to efficiently generate realistic, high-resolution SAR and ISAR images of any 3D model in a Standard Tessellation Language (STL) format. The program is currently available in Windows' platforms, RadarPixel's layout is showcased in Figures ~\ref{radarPixel}.a, where we can observe most of the simulator's available functions, and how the STL model looks once it is inside the simulation.

In order to run a simulation in any given aircraft we have to input the angle of elevation in which the SAR happened to receive the information from a passing aircraft, also we need the distance from the radar's position to the aircraft, so using basic trigonometric operations to get the height of the plane using the elevation angle and a fixed distance from the radar to the target we get the distance from the ground to the target (height), the final distance of the aircraft is calculated using the root of the sum of the squares of the distance and height, then we calculate the velocity of the target using the final distance times achieved at the end of the simulation; after the simulation has completed its motion compensation calculations we can observe a sample image in Figure ~\ref{radarPixel}.b and Figures ~\ref{fig:MSTARvRadarPixel1} and ~\ref{fig:MSTARvRadarPixel2} for the Slicy.

The simulation also includes a preset for the amount of noise in the environment; signal noise is an internal source of random variations in the signal, which is generated by all electronic components. Reflected signals decline rapidly as distance increases, so noise introduces a radar range limitation; this noise is generated in the simulator using the Gaussian noise function available in MATLAB. Noise typically appears as random variations superimposed on the desired echo signal received by the radar receiver. The lower the power of the desired signal, the more difficult it is to discern it from the noise. The stealth aircraft rely on this noise to hide from any radar signals since the stealth material that they carry impedes the radar waves from bouncing correctly off of the aircraft's body.\\

\subsection{Validation}

\begin{figure}
  \includegraphics[width=\linewidth]{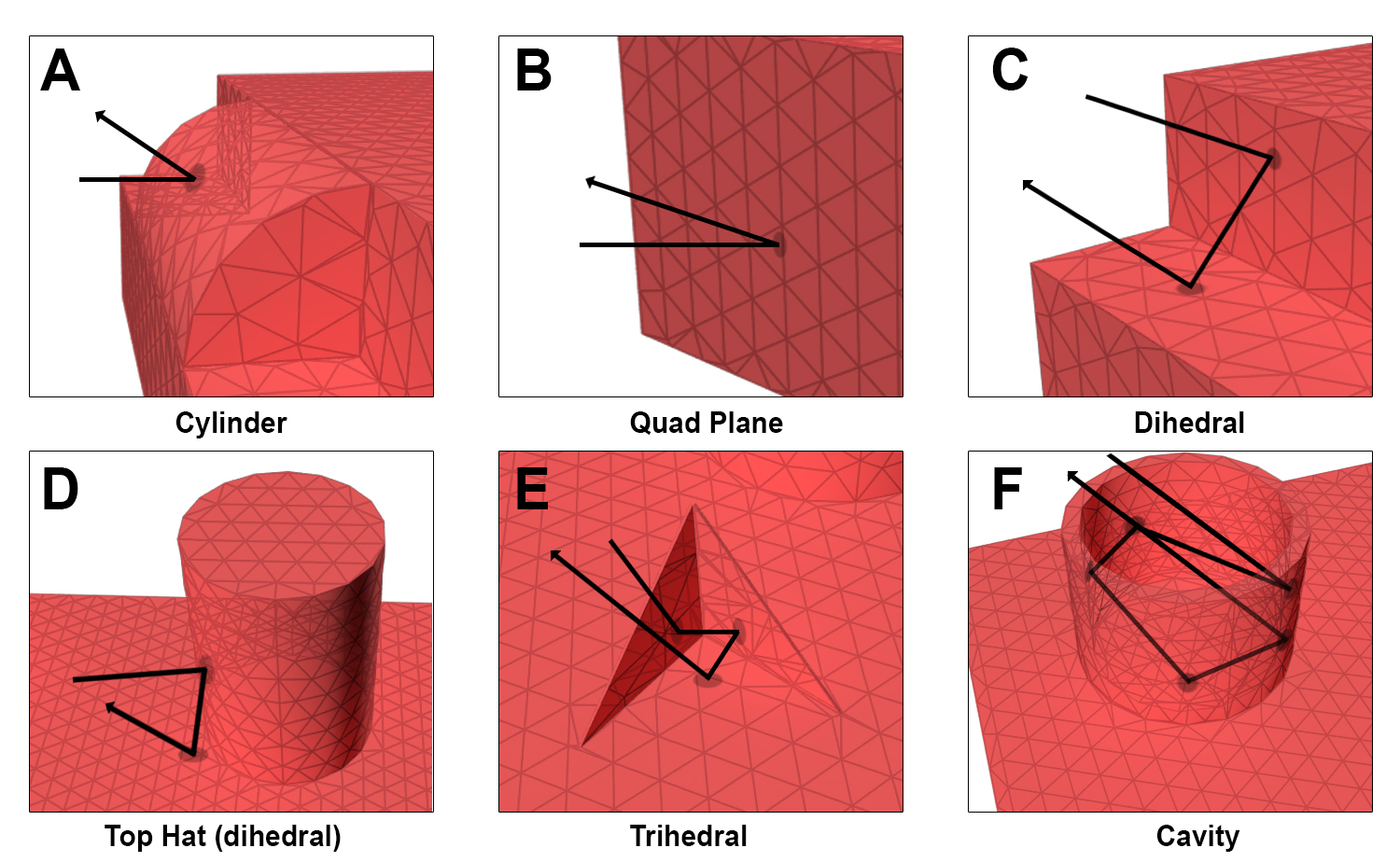}
  \caption{Scattering primitives available in the Slicy target.}
  \label{fig:Primitives}
\end{figure}

"The Slicy target is a precisely designed and machined engineering test target... to allow Image Understanding developers the ability to validate the functionality of their algorithm with a simple know target" \cite{4}; basically the Slicy is built test the SAR capabilities, since it has every possible scattering primitive that could be encountered in an object. In Figure ~\ref{fig:Primitives} we can observe the different scattering techniques that can be a applied to every section of the Slicy, for example, single bounce (flat-plate), double bounce (dihedral), triple bounce (trihedral), edge diffraction (cylinder and top hat), cavity (hollow cylinder) and shadowing (obstructions between parts on the target).

\begin{figure}
  \includegraphics[width=\linewidth]{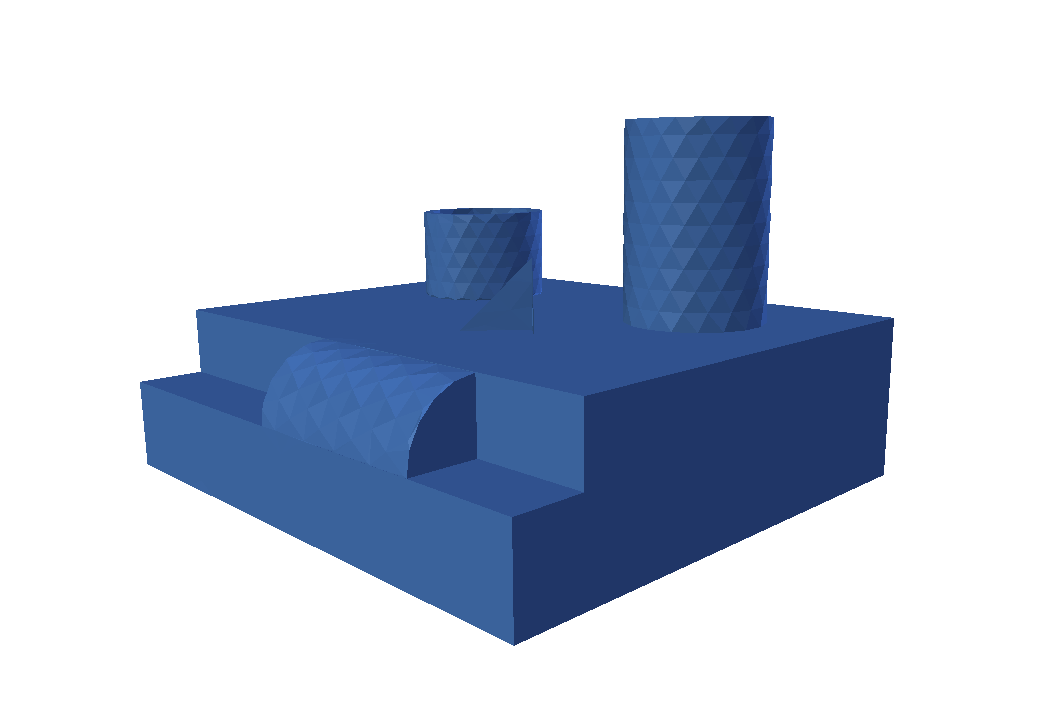}
  \caption{Slicy model.}
  \label{fig:Slicy_45deg}
\end{figure}

\begin{figure*}
  \centering\includegraphics[width=\linewidth]{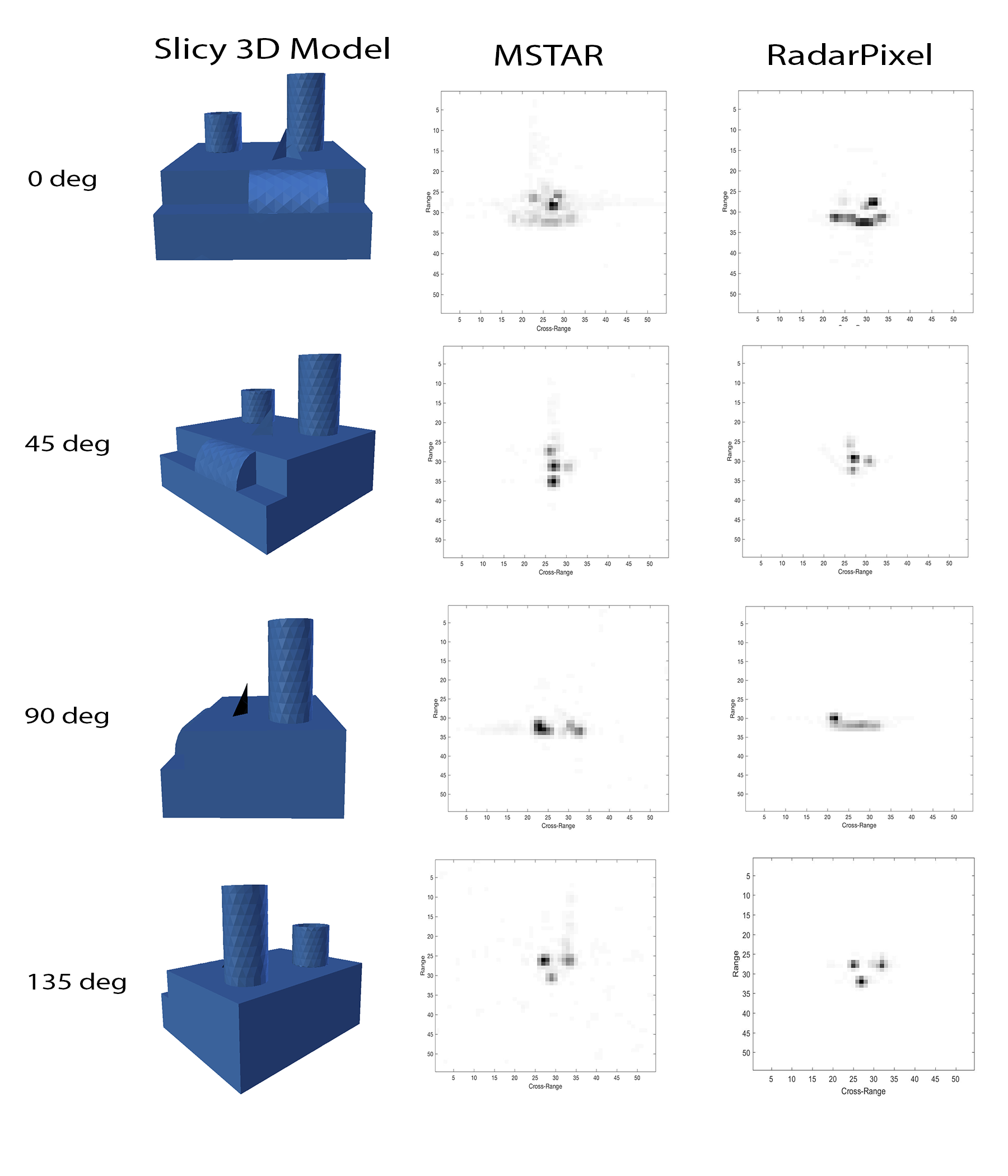}
  \caption{Comparison between theMSTAR data from the real Slicy and the RadarPixel's simulated images at 15-degrees of elevation angle.}
  \label{fig:MSTARvRadarPixel1}g
\end{figure*}

\begin{figure*}
  \centering\includegraphics[width=\linewidth]{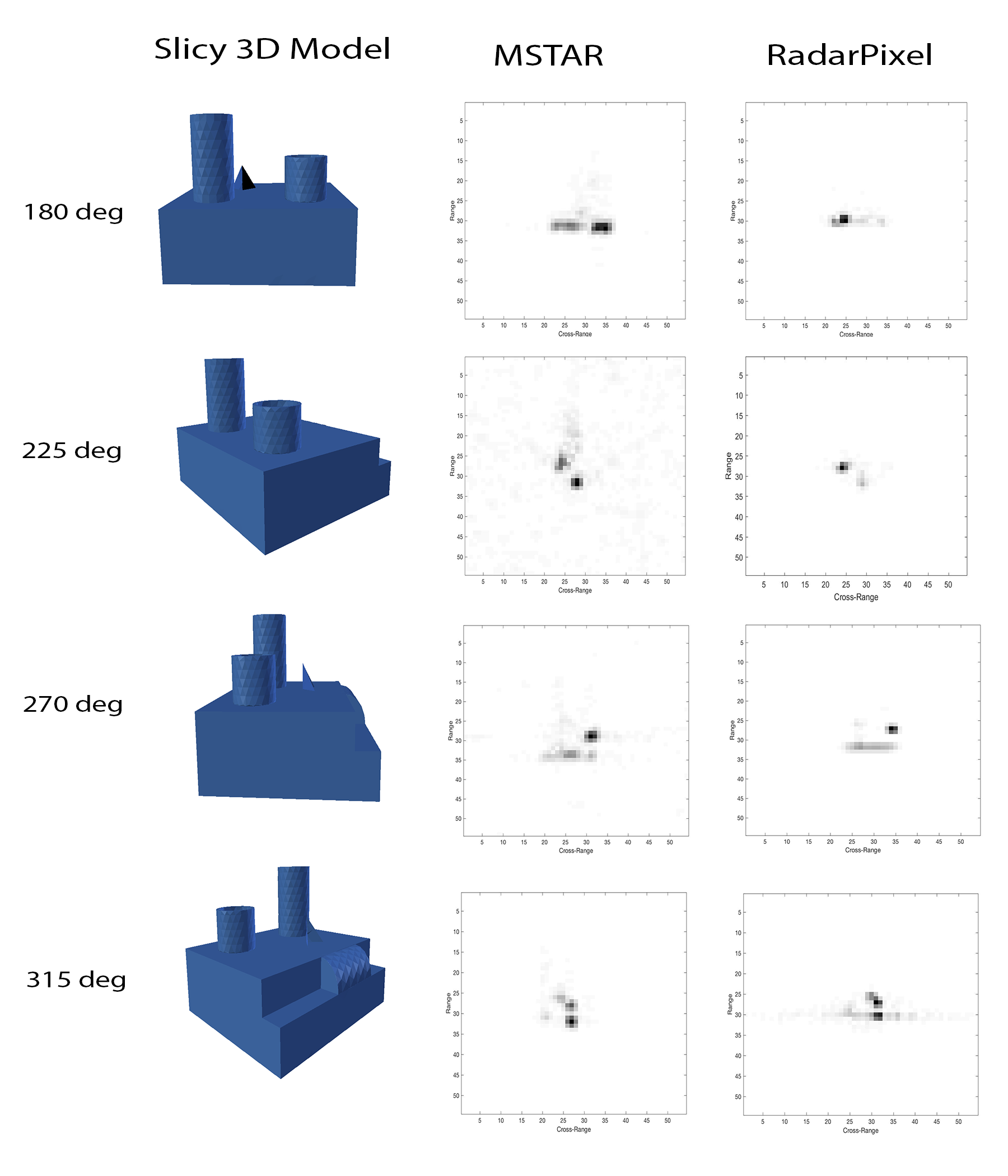}
  \caption{Comparison between theMSTAR data from the real Slicy and the RadarPixel's simulated images at 15-degrees of elevation angle.}
  \label{fig:MSTARvRadarPixel2}
\end{figure*}

The parameters used for the validation of the simulation were directly inspired from \cite{4} where they tested a real SAR with a Slicy made out of concrete and it was 2.445 meters wide, 2.75 meters long, and 0.765 meters tall (only for the rectangular box) + 0.915 meters extra from the tallest cylinder. We obtained from the simulator 2 degrees of elevation (15 and 30) and 8 degrees of rotation (0, 45, 90, 135, 180, 225, 270, 315). Our Slicy model \cite{26} (Figure ~\ref{fig:Slicy_45deg}) contained 6400 total faces and we can observe in Figure~\ref{fig:MSTARvRadarPixel1} and ~\ref{fig:MSTARvRadarPixel2} our results nearly match the real world test of every angle, although if we introduced more total faces the images will begin to be even more similar. Thus we can conclude that our simulator is accurately creating SAR images, that match real-world scenarios.\\

\section{Methods}

\subsection{Convolutional Neural Network}
\begin{figure}[h!]
  \includegraphics[width=\linewidth,height=5.0in]{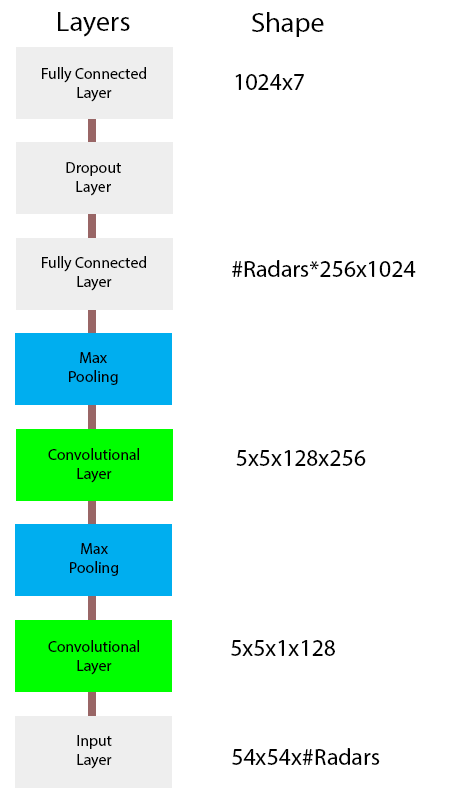}
  \caption{Convolutional Neural Network Architecture.}
  \label{CNN}
\end{figure}

To test our hypothesis that using Multi Mono-static for automatic target recognition will increase the accuracy for image classification, we constructed a CNN. A CNN is a type of deep net that helps analyze image data, based on a feed-forward artificial neural network \cite{1}. For data generation, we used the same degrees of elevation as the Slicy model, but we generated 360 images per elevation in order to get a full 360° view of the aircraft model. Afterwards, we modified the level of noise we allowed the simulator to generate. Noise in electronic components is caused by different factors such as weather and lighting.
The architecture of the network is inspired by the basic MNIST but we used a 54x54 image size with complex number matrix instead of pixels from an image, this gives the network a lot more information about the image. Our first layer of convolution will compute 128 features for each 5x5 patch and depending on the number of radars in the simulation is the number of channels that the input layer will have. The max pooling layer will reduce the matrix size to 27x27, then the second convolutional layer will get 256 features every 5x5 patch from the 27x27 input. Finally, a last max pooling layer will return a 14x14 matrix and we can add a fully connected layer with 1024 neurons to allow processing on the entire matrix. During the dropout layer, we define a probability that the output of a neuron is kept (On during training and off during testing). At last, a resulting in output a number between 0 and 7 where each number is defined by a specific aircraft, the order of the aircraft can be seen in the Tables 1-5. See Figure ~\ref{CNN} to observe the flow of our networks architecture \cite{27}.

\subsection{Multiple Mono-static Radar}
Mono-static Radar images are generated using a single radar that sends a signal and receives the bounced signals from the desired target (see Figure ~\ref{radars}.a); a Multiple Mono-static radar is that which comprises multiple radar-generated images of the mono-static radar, using this method we increased our detection accuracy significantly since the network was getting a complete view of the aircraft (see Figure ~\ref{radars}.b).The parameters used to fulfill the requirements for the multiple Mono-Static sets of images were given by the number of radars that we chose to test, for example, for a 4 radar setting we get the degrees of rotation for 0, 89, 179, 269 then increasing the numbers by one giving us a total of 90 sets for a given degree of elevation so having 2 degrees of elevation we get a total of 180 sets of images to work on that given number of multiple Mono-Static radars.\\

\subsection{SARS}
The datasets created for testing our simulation comes from the aircraft fighter jet 3D models selected from US aircraft: F15 and F16, Chinese Aircraft: J11 and J15, and Russian aircraft MIG29 and MIG35. The data is composed of a combination of elevation and rotation angles (the aircraft can be seen in Figure ~\ref{aircraft}), and a distance is given as a parameter to the radar simulator that represents the position of elevation of the radar simulator from where it will be pointing towards the target. The velocity and other parameters are then computed by the simulator software. The result is a vector-image transformed out of the radar simulator output features and their class label. The SAR simulator output is based on the scattering primitives and a number of reflections returned by the target model. The intensity of the output image features is dependent on a number of reflections of the model. Meshes and scattering primitive geometric shapes form a very important aspect of retrieving reliable data from each target simulated.
The simulator’s range of generated noise in the image is from 201 dB, which is considered as no noise, and 0 dB, mainly noise in the image. We generated images for 201 dB, 200 dB, 150 dB, 100 dB, and 50 dB with Gaussian noise signal to noise ratio; for each noise level, we generated 5040 images for each aircraft we had available, thus giving a grand total of 25,200 images. After testing all the generated data with our network we observed that 201 dB was basically generating the same imaging as with 200 dB thus 201 dB was discarded from results, and we observed 50 dB had too much noise that it would not be possible to run it successfully through a network, therefore 50 dB was not considered; after discarding all the useless images we are left with 15, 120 images for training and testing.\\

\begin{figure}[h!]
  \includegraphics[width=\linewidth]{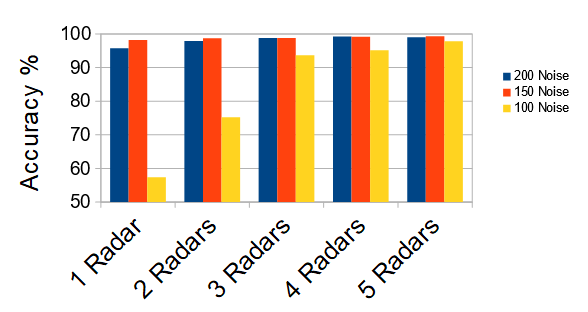}
  \caption{Accuracy for 7 aircraft models at 200, 150, and 100 noise levels.}
  \label{graph}
\end{figure}

\section{Experimental Results}
With the usage of a 10-fold cross-validation to train and test the neural network, we can verify (as shown in Figure ~\ref{graph}) the results of the most significant tests we ran through our CNN. We observed that the number of radars does not affect directly to the accuracy of the models when the noise level is 200 dB or 150 dB. That indicates that in an environment with low or normal noise levels, the CNN will classify the aircraft with nearly a hundred percent accuracy. but when the radars are receiving noise. Similarly, in Table 1 we can observe that the accuracy per class when using only a Mono-static approach will be kept high when no noise is introduced in the radar, but it will significantly be decreased if we don’t use multi Mono-static approach, and Tables 2-5 show the accuracy for every aircraft we tested separated by the number of radars used at each instance.\\

\begin{table}[h!]
\centering
\caption{Accuracy per class for Mono-Static}
\label{1 Radarl}
\begin{tabular}{|c|c|c|c|}
\hline
Model  & 200 dB & 150 dB & 100 dB \\ \hline
F15    & 99.16$\pm$1.13   & 98.59$\pm$0.74   & 56.38$\pm$4.27   \\ \hline
F16    & 98.88$\pm$1.53   & 98.04$\pm$1.1   & 62.63$\pm$11.12   \\ \hline
J11    & 95.83$\pm$1.86   & 91.70$\pm$2.82   & 42.91$\pm$5.17   \\ \hline
J15    & 97.08$\pm$2.40   & 90.42$\pm$2.0   & 50.83$\pm$7.4   \\ \hline
MIG29  & 98.33$\pm$1.76   & 96.40$\pm$1.03   & 56.11$\pm$6.24   \\ \hline
MIG35  & 98.61$\pm$4.16   & 96.95$\pm$1.55   & 55.97$\pm$6.18   \\ \hline
EF2000 & 99.16$\pm$2.15   & 98.02$\pm$1.02   & 76.80$\pm$4.98   \\ \hline
\end{tabular}
\end{table}

\begin{table}[h!]
\centering
\caption{Accuracy per class with 2 Radars}
\label{2 Radars}
\begin{center}
\begin{tabular}{|c|c|c|c|}
\hline
Model  & 200 dB & 150 dB & 100 dB \\ \hline
F15    & 99.44$\pm$1.10   & 100$\pm$0.0   & 78.05$\pm$7.26   \\ \hline
F16    & 100$\pm$0.0    & 99.72$\pm$0.93   & 79.16$\pm$8.18   \\ \hline
J11    & 95.55$\pm$3.99   & 96.94$\pm$2.15   & 64.72$\pm$6.14   \\ \hline
J15    & 92.77$\pm$5.13   & 95.83$\pm$3.09   & 66.38$\pm$5.88   \\ \hline
MIG29  & 98.33$\pm$2.56   & 99.16$\pm$1.44   & 77.50$\pm$7.98   \\ \hline
MIG35  & 99.16$\pm$1.35   & 99.16$\pm$1.49   & 75.55$\pm$6.83   \\ \hline
EF2000 & 99.72$\pm$0.85   & 100$\pm$0.0   & 91.94$\pm$4.25   \\ \hline
\end{tabular}
\end{center}
\end{table}

\begin{table}[h!]
\centering
\caption{Accuracy per class with 3 Radars}
\label{3 Radars}
\begin{tabular}{|c|c|c|c|}
\hline
Model  & 200 dB & 150 dB & 100 dB \\ \hline
F15    & 100  $\pm$0.0    & 100  $\pm$0.0   & 98.33$\pm$2.02   \\ \hline
F16    & 99.16$\pm$1.78   & 100  $\pm$0.0   & 95.83$\pm$3.82   \\ \hline
J11    & 99.66$\pm$3.04   & 98.75$\pm$2.99   & 91.25$\pm$6.82   \\ \hline
J15    & 97.08$\pm$4.16   & 97.08$\pm$4.95   & 90.83$\pm$6.35   \\ \hline
MIG29  & 100  $\pm$0.0    & 99.58$\pm$1.22   & 98.75$\pm$2.14   \\ \hline
MIG35  & 99.16$\pm$1.84   & 98.75$\pm$3.16   & 96.66$\pm$3.53   \\ \hline
EF2000 & 99.58$\pm$1.05   & 100  $\pm$0.0   & 97.91$\pm$2.58   \\ \hline
\end{tabular}
\end{table}

\begin{table}[h!]
\centering
\caption{Accuracy per class with 4 Radars}
\label{4 Radarsl}
\begin{tabular}{|c|c|c|c|}
\hline
Model  & 200 dB & 150 dB & 100 dB \\ \hline
F15    & 100   $\pm$0.0   & 100  $\pm$0.0   & 97.77$\pm$2.37   \\ \hline
F16    & 100   $\pm$0.0   & 100    $\pm$0.0   & 95.55$\pm$5.47   \\ \hline
J11    & 96.11$\pm$6.51  & 98.33$\pm$2.59   & 92.77$\pm$6.78   \\ \hline
J15    & 98.33$\pm$3.0   & 98.33$\pm$4.27   & 82.77$\pm$12.66   \\ \hline
MIG29  & 100 $\pm$0.0   & 100    $\pm$0.0   & 94.44$\pm$5.26   \\ \hline
MIG35  & 100 $\pm$0.0   & 98.33 $\pm$3.06   & 91.66$\pm$6.52   \\ \hline
EF2000 & 100$\pm$0.0   & 100     $\pm$0.0   & 97.22$\pm$2.7   \\ \hline
\end{tabular}
\end{table}

\begin{table}[h!]
\centering
\caption{Accuracy per class with 5 Radars}
\label{5 Radars}
\begin{tabular}{|c|c|c|c|}
\hline
Model  & 200 dB & 150 dB & 100 dB \\ \hline
F15    & 100  $\pm$0.0    & 100  $\pm$0.0   & 99.30$\pm$1.58   \\ \hline
F16    & 100  $\pm$0.0    & 100  $\pm$0.0   & 100$\pm$0.0     \\ \hline
J11    & 97.18$\pm$6.41   & 98.58$\pm$2.04   & 95.80$\pm$5.72   \\ \hline
J15    & 97.16$\pm$4.83   & 95.83$\pm$6.24   & 92.19$\pm$6.37   \\ \hline
MIG29  & 99.30$\pm$1.86   & 99.30$\pm$2.11   & 99.29$\pm$2.26   \\ \hline
MIG35  & 100  $\pm$0.0    & 97.90$\pm$5.27   & 97.91$\pm$4.99   \\ \hline
EF2000 & 99.30$\pm$2.43   & 100  $\pm$0.0   & 100$\pm$0.0     \\ \hline
\end{tabular}
\end{table}

\section{Conclusion}
In this paper we presented a new simulation software with a novel approach to classify military aircraft using multiple  Mono-static radars to generate an array of pictures around the target at any given moment and feeding the array as a matrix to our CNN, we can observe from the results and Tables 1-5 that the best number of radars to have in such array has to be greater than 4 since it will give results that are nearly as accurate as radars in the best climatic and lit conditions.

\section{Discussion}
The original scope of this paper included military stealth aircraft like F35, J20, T50, etc., but due to the little trustworthy information for the stealth coating and material these aircraft use to absorb the electromagnetic waves from the SAR/ISAR, we could not include those results in this paper, as well as noise below 50 dB which will cause the images to be too randomized leading to inaccurate training within the size of the current number of images in our dataset.\\
Although the results of this paper proved that having more radars will increase the accuracy considerably we are aware that SAR/ISAR equipment is expensive and having several of them could be nearly impossible, so in future work for these study we will utilize a Bi-static method that should lead to the same conclusion as our current one, but with Bi-static only one sender/receiver is necessary for the array and the rest will be only receiving the bounced electromagnetic waves from the target.


\section{Acknowledgment}
We gratefully acknowledge the support of NVIDIA Corporation with the donation of the Titan X Pascal GPU used for this research.



%

\end{document}